\pdfoutput=1
\documentclass[11pt]{article}

\usepackage[T1]{fontenc}
\usepackage[utf8]{inputenc}
\usepackage{times}
\usepackage{microtype}

\usepackage{inconsolata}
\usepackage{amsmath}

\usepackage[preprint]{acl}

\usepackage{hyperref}
\usepackage{graphicx}
\usepackage{xcolor}
\usepackage{epsfig,url,svg}
\usepackage{float}
\usepackage{comment}
\usepackage{framed}
\usepackage{booktabs}
\usepackage{rotating}
\usepackage{xcolor,colortbl}
\usepackage{enumitem}

\newlist{inlinelistone}{enumerate*}{1}
\setlist*[inlinelistone,1]{%
  label=(\textbf{\arabic*)},
}
\newlist{inlinelisttwo}{enumerate*}{1}
\setlist*[inlinelisttwo,1]{%
  label=(\textbf{\roman*)},
}

\usepackage{enumitem,amssymb}
\usepackage{multirow}

\author{
    Evangelia Gogoulou$^{1,3}$~~~~Shorouq Zahra$^{1}$~~~~Liane Guillou$^{4}$~~~~Luise D{\"u}rlich$^{1,2}$~~~~Joakim Nivre$^{1,2}$ \\[2mm]
    $^{1}$RISE Research Institutes of Sweden, Department of Computer Science \\
    $^{2}$Uppsala University, Department of Linguistics and Philology \\
    $^{3}$KTH Royal Institute of Technology, Division of Software and Computer Systems \\
    $^{4}$University of Edinburgh, School of Informatics \\[2mm]
    {\normalsize \texttt{\{evangelia.gogoulou, shorouq.zahra, luise.durlich\}@ri.se}} \\
    {\normalsize \texttt{liane.guillou@ed.ac.uk~~~~joakim.nivre@lingfil.uu.se}}
}

\begin{document}
\title{Can LLMs Detect Intrinsic Hallucinations \\[1mm] in Paraphrasing and Machine Translation?}
% \title{Intrinsic Hallucination Detection in Paraphrasing and Machine Translation}
%\title{Detecting Intrinsic Hallucinations with Large Language Models}

%\title{Hallucination Detection with Large Language Models}

\maketitle              % typeset the header of the contribution

\begin{abstract}

%The phenomenon of hallucinations, referring to the specific case when the model output is unfaithful to the input, is a frequently occurring problem in LLMs. 
A frequently observed problem with LLMs is their tendency to generate output that is nonsensical, illogical, or factually incorrect, often referred to broadly as ``hallucination''.   
Building on the recently proposed HalluciGen task for hallucination detection and generation, we evaluate a suite of open-access LLMs on their ability to detect intrinsic hallucinations
%hallucinations. We focus on
%We focus on a special case of 
%intrinsic hallucination, namely generating output that is not entailed by the input 
in two conditional generation tasks: translation and paraphrasing.
%in their input, investigating how the model performance changes between languages. 
We study how model performance varies across tasks %(paraphrase or translation) and across 
and language and we investigate the impact of %formulation, 
model size, instruction tuning, and prompt choice.
%The additional factors of prompt input, instruction tuning and model size are also analysed.
% We find that the level of performance differs between LLMs, but each model overall has a consistent performance across languages in each task. For model size and instruction tuning, the effect varies between model families, with a consistent positive effect only for the Llama-3 models, and that language affects the performance more than prompt formulation.
We find that performance varies across models but is consistent across prompts. % Further, the effects of model size and instruction tuning vary across model families.
%and for prompts it appears that language has a greater effect on performance than formulation. 
%We find that LLMs are able to detect hallucinations with performance varying between NLP tasks (paraphrase or translation) and languages  %prompt formulation does not have a consistent effect on the performance. For model size and instruction tuning, we find that the effect varies between model families, with a consistent positive effect only for the Llama-3 models.
%, the language (pair), and the prompt formulation. 
Finally, we find that NLI models perform comparably well, suggesting that LLM-based detectors are not the only viable option for this specific task. %of hallucination detection. 
\end{abstract}

\section{Introduction}

The introduction of large language models (LLMs) has revolutionised the field of natural language processing (NLP). State-of-the-art LLMs have demonstrated excellent language generation capabilities %and suitability 
in 
%broad applications %such as 
conversational AI \cite{zhao2024surveylargelanguagemodels},
%They also exhibit 
as well as strong performance on more specific NLP tasks like summarisation \cite{pu2023summarizationalmostdead}, open-domain question answering \cite{kamalloo-etal-2023-evaluating}, sentiment analysis \cite{zhang-etal-2024-sentiment}, and machine translation \cite{kocmi-etal-2023-findings}. Despite this success, LLMs are prone to producing output that is fluent and grammatical but semantically inadequate or factually incorrect,
%nevertheless fluent but inaccurate outputs that are not faithful to the given input, 
a phenomenon broadly referred to within the NLP community as ``hallucination''. 
The impact of hallucinations by LLMs may be severe in downstream applications where accurate output is mission critical, or %for example 
where hallucination leads to erroneous decisions %being made 
with negative consequences that directly impact humans e.g. in the medical or legal domain. In many cases, it may be infeasible to have a human in the loop, or it may be difficult for humans to identify hallucinations, which motivates the need for automated methods for detection and evaluation.% evaluating and detecting hallucinations. 

% Liane: add this back into the paper when we expand it to include generation
%Whilst there has been a strong focus on the \textit{problem} of hallucination, it may also be seen as a \textit{feature} of LLMs to be exploited with positive effect \textcolor{red}{XXXX - add citation - XXXX}. For example, it may be useful for data augmentation, and in particular for the generation of adversarial examples for training and evaluation purposes. This motivates the inclusion of a generation step in the task setup. 

% Liane: merge this back into the paper when we expand it to include generation
%In this paper we explore both of these directions. We aim to discover...

In this paper, we aim to discover whether LLMs can be used to detect hallucinated content, focusing on a special case of what \citet{hallucination_llm_survey}
call intrinsic hallucinations, that is, cases where the output is deficient with respect to a particular input and where the deficiency can be detected given only the input and output.\footnote{This in contrast to extrinsic hallucinations, where additional information such as world knowledge is required to detect the deficiency.} More precisely, for the tasks of paraphrasing and machine translation, we define a hallucination to be an output, or hypothesis, that is not entailed by the input, or source.%\footnote{This is a generalization of the definition in \citet{hallucination_llm_survey}, where hallucinations are required to contradict the source.} %Taking this a step further, we also explore the viability of using LLMs as a judge in a cross-evaluation setting, where one LLM is used to evaluate the output of another \cite{Zheng2023, manakul2023selfcheckgpt, saunders2022self}.

We build upon %our
previous work from the ELOQUENT Lab at CLEF 2024 \cite{durlich2024overview}, where the HalluciGen task asked participants to apply LLMs %to the task of 
in detecting and generating hallucinations. %(HalluciGen task). %We build upon our previous work from the ELOQUENT Lab at CLEF 2024 \cite{durlich2024overview}, specifically the HalluciGen task, where we asked participants to apply LLMs to the task of detecting and generating hallucinations.
We extend the work from the shared task with a series of experiments in prompting open-access LLMs to detect hallucinations.
%for the task of hallucination detection. %Hallucination detection 
It is framed as a contrastive challenge task: given a source sentence, and a pair of hypotheses, the model should detect which one contains a hallucination. 

We evaluate a range of open-access LLMs on hallucination detection in paraphrase generation and translation, as defined in the HalluciGen task \cite{durlich2024overview}. Through a systematic investigation of model performance on the hallucination detection task, we address
the following 
%research 
questions: 
        \begin{itemize}[noitemsep,topsep=2pt,leftmargin=7mm]
        \item How does model performance differ across target languages?
        \item Does increased model parameter size improve performance?
        \item Does instruction tuning improve performance? 
        \item Does the language and formulation of the prompt matter?
        \end{itemize}

\section{Background and Related Work}
Two concepts that are often used to characterize different types of hallucinations are \emph{faithfulness} and \emph{factuality}. Faithfulness means being consistent with a given source or input, and has long been used as an evaluation criterion in conditional generation tasks like machine translation; a faithfulness hallucination is therefore any output that lacks such consistency, regardless of whether it is factually correct. By contrast, factuality means corresponding to real-world knowledge, and a factuality hallucination is therefore any output that makes a false claim, regardless of context and input. A related distinction is made between \emph{intrinsic} and \emph{extrinsic} hallucinations, where the former can be detected by only considering the input and output of a system, while the latter requires access to additional information %such as general world knowledge 
\citep{hallucination_llm_survey}.

Prior work has mostly focused on building systems to detect 
factuality hallucinations. For example, \citet{li-etal-2023-halueval} introduce a benchmark targeting cases of factual hallucinations in the context of question-answering, knowledge-grounded dialogue, and summarisation. %We instead focus on faithfulness and intrinsic hallucinations, investigating the capabilities of LLMs to detect hallucinations with (only) the model input as reference. 
Aside from the HalluciGen task
the closest work to ours is the SHROOM shared task \cite{mickus2024semeval} from SEMEVAL 2024. SHROOM defines hallucinations as cases when the hypothesis cannot be inferred from its semantic reference. Despite the similarity with our definition, there is a significant difference in how the hallucinations are constructed. In SHROOM they are generated by models prompted to solve the specific task scenario, whilst we mostly construct hallucinations manually based on specific categories of %entity and semantic 
errors; by switching gender, negation, or tense, replacing words with their antonyms, by substituting named entities, numbers, dates, and currencies, and by making superfluous additions. %However, for some cases, we also rely on LLMs to generate %real 
%hallucinated responses to downstream scenarios. These do not fit into any of the above categories and are referred to as natural hallucinations. 
%This breakdown of the hallucination types allows us to systematically study the model performance across different types of hallucinations. %Unlike the framing of SHROOM task which requires a binary response, our task is contrastive. %
The two tasks also differ in terms of their coverage of NLP tasks and target languages. SHROOM includes the additional task of definition modeling; HalluciGen covers an extra language for paraphrase but has limited coverage for machine translation.

%Previous work on %%investigating model performance on 
%hallucination detection is limited. 
There is limited evidence so far on the effectiveness of using LLMs for detecting hallucinations. 
\citet{li-etal-2023-halueval} find that LLMs, including Llama2 and ChatGPT, perform poorly on the task of identifying hallucinations that have been generated by LLMs to be factually incorrect, in English question-answering and summarisation.
According to the HalluciGen task results \cite{durlich2024overview}, GPT-4 and LLM majority voting approaches outperform smaller English-centric models such as Llama3-8b and Gemma-7b.
Similar conclusions emerge from SHROOM, where submissions based on GPT-4 or model ensembling exhibit the strongest performance. Model fine-tuning on SHROOM training data is another successful approach.   %Even though our work focuses on contexts with shorter length, we aim to study hallucinations in longer contexts in the future. 

NLI models have also shown promising abilities at detecting hallucinations. \citet{manakul2023selfcheckgpt} propose SelfCheckGPT to detect sentence-level hallucinations 
%using a variety of approaches, including 
using generative LLM prompting, LLM probabilities, and NLI models. Interestingly, their experimental results show that LLM prompting  %``SelfCheckGPT with Prompt''
outperforms the %``SelfCheckGPT with NLI''%
NLI-based method only by a small margin, and both outperform all other SelfCheckGPT methods and baselines. % on sentence-level hallucination detection. 

Additionally, NLI-based methods %methods 
%systems
%metrics
%have 
% have also shown 
yield
promising results for high-resource languages in multilingual setups, often outperforming other lexical metrics (like ROUGE), especially for %``verifiable hallucinations'' 
intrinsic hallucinations where the hypothesis would clearly contradict the source \cite{Kang2024ComparingHD}.

The ability of NLI models to detect intrinsic hallucinations is arguably unsurprising as they must ``handle phenomena like lexical entailment, quantification, coreference, tense, belief, modality, and lexical and syntactic ambiguity'' \citep{williams-etal-2018-broad} to successfully predict entailment, contradiction, and neutral relations between sentence pairs. 

\section{Dataset Description}
%Our dataset covers two scenarios:
The HalluciGen detection task \cite{durlich2024overview} covers the two following scenarios:
\begin{itemize}[topsep=2pt,noitemsep]
    \item \textbf{Paraphrase Generation}: The model is presented with two possible paraphrases of a given source sentence in English (en) and Swedish (sv).
    \item \textbf{Machine Translation}: Given a sentence in a source language, the model is presented with two possible translations in the target language; %We include two language pairs: English-German 
    English-German (en$\Leftrightarrow$de) and English-French (en$\Leftrightarrow$fr), in both translation directions.
\end{itemize}
Each example in the dataset consists of a source sentence ($src$), a good hypothesis ($hyp+$), and an incorrect hypothesis containing an intrinsic hallucination ($hyp-$). The criterion for a hypothesis to contain such a hallucination is that it is not entailed by the source sentence, which in turn means that it must contain some additional or contradictory information with respect to the source. This may be due to additions, substitutions, negations, or other phenomena that break the inference relation.  
% (and therefore must contain instances of addition, substitution, and/or alterations that make it incorrect in respect to its source).
Note that %our
this definition is a relaxation of the definition in \citet{hallucination_llm_survey}, where intrinsic hallucinations are required to explicitly contradict the source. Note also that a hypothesis that does not entail the source sentence is not considered a hallucination, despite being an imperfect paraphrase/translation, as long as it is still entailed by the source. For example, if the source is ``it is cold and wet'', then ``it is cold and windy'' and ``it is not cold and wet'' are both considered hallucinations, but ``it is cold'' is not.

Each hallucinated hypothesis belongs to one of these eleven categories, defined by the type of error or addition that breaks the entailment relation %from the source: 
addition, named-entity, number, conversion, date, gender, pronoun, antonym, tense, negation, natural. 
The last category refers to hallucinated responses by LLMs that did not fit into any of the other above categories.
%For those cases, we rely on LLMs to generate 
%hallucinated responses to downstream scenarios. These do not fit into any of the above categories and are referred to as ``natural'' hallucinations.
%(i.e. a hallucination that does not fit into any of the other categories). 
Examples of each hallucination category for the paraphrase task can be found in Table~\ref{tab:examples} in Appendix~ \ref{sec:hallucination_examples}, and the frequency statistics of the hallucination categories in Appendix~\ref{sec:category_statistics}.
%In addition examples in the dataset include hallucinations for eleven categories: 
%We release all datasets on Huggingface.\footnote{\url{https://link-to-be-provided-for-camera-ready}} 
All datasets are available on Huggingface.\footnote{\url{https://huggingface.co/datasets/Eloquent/HalluciGen-PG} (paraphrase) and \url{https://huggingface.co/datasets/Eloquent/HalluciGen-Translation} (translation)} The dataset creation process for the translation and paraphrase scenarios is summarised below and described in full in \citet{durlich2024overview}.

\subsection{Paraphrase Generation}
%For the English dataset, we sampled $138$ examples from the SHROOM training set for the paraphrase generation subtask \cite{mickus2024semeval}. 
The English dataset consists of $138$ examples from the SHROOM training set for the paraphrase generation subtask \cite{mickus2024semeval}. For the Swedish dataset, %we used 
$139$ examples from the SweParaphrase test data were used \cite{berdicevskis-etal-2023-superlim}, consisting of sentence pairs together with their degree of semantic similarity, and the Swedish part of the Finnish Paraphrase Corpus \cite{kanerva-etal-2021-finnish}, which consists of paraphrase hypothesis pairs and a label indicating the degree of paraphrase relation. %We selected examples with 
The selected examples have the highest 
%degree of 
similarity (SweParaphrase), or are paraphrase equivalents (Finnish Paraphrase Corpus). %prioritising those with the highest degree of semantic similarity. 

%We then used 
Mixtral-8x7B-instruct \cite{jiang2024mixtral} and GPT-SW3-6.7B-instruct \cite{ekgren-etal-2024-gpt} were used to automatically generate a paraphrase hypothesis for the first sentence of each pair, after which all examples were manually annotated in two steps. The annotators first determined whether the generated hypothesis is an intrinsic hallucination with respect to the source (see Appendix~\ref{app:ann_guidelines} 
%for the annotation guidelines given to the annotators
). %, given the definition of hallucination: \textit{a hypothesis contains a hallucination if it is not supported by the source}. 
Then for those hypotheses not marked as hallucinations, the annotators manually constructed a hallucination based on one of the ten first categories (i.e.\ excluding natural hallucinations). %in Table~\ref{tab:examples}%,
The hypotheses marked as hallucinations were assigned to one 
%of those categories
type, or the natural type if they did not correspond to any specific hallucination phenomenon.
%none of the other categories was applicable.

The test set for each language consists of $119$ examples, %and there are additionally 
with $16$ additional trial examples for English and $20$ %trial examples 
for Swedish. 
We computed the inter-annotator agreement on 
the binary classification (hallucination or not) by three annotators, measured using Krippendorff's alpha, shows high agreement: $0.90$ for English and $0.88$ for Swedish. The annotation guidelines are presented in Appendix ~\ref{app:ann_guidelines}.

\subsection{Machine Translation}

%We 
\citet{durlich2024overview} leveraged ACES \cite{amrhein_aces_2022}, a contrastive challenge set for evaluating machine translation metric performance on a range of translation accuracy errors. ACES examples consist of a source sentence, a pair of good/incorrect translation hypotheses, a reference translation, and a label denoting the error phenomenon in the incorrect translation. As ACES already contains examples for en$\Leftrightarrow$fr and en$\Leftrightarrow$de for most of the hallucination categories (except tense and negation) %we sampled 
the majority of %our 
the HalluciGen
dataset examples were sampled directly. For the tense and negation categories, %we constructed 
new examples were constructed using the PAWS-X dataset \cite{yang-etal-2019-paws} of adversarial paraphrases. 

%We sampled $100$ test examples for each language direction 
For each language direction, $100$ test examples were sampled from the categories of ACES aiming for a uniform distribution across these categories as much as possible. Additionally, %we 
$10$ trial examples were selected for each language direction.

\section{Experimental Setup}

\subsection{Models}
We evaluate a range of different model families, which differ in the type and amount of pre-training language data. From each family, we select multiple model variants that differ in model size and/or presence of instruction tuning. This enables the systematic study of those two factors in relation to the ability of the model to detect hallucinations. We select a number of variants from the \textbf{Llama3} \cite{dubey2024llama}, \textbf{Mixtral} \cite{jiang2024mixtral}, \textbf{EuroLLM}, and \textbf{GPT-SW3} \cite{ekgren-etal-2024-gpt} model families. The full list of models is found in Appendix \ref{sec:model-repos}. The GPT-SW3 models are evaluated only in the paraphrase scenario, while the rest are used for both scenarios. %The reported presence of each target language during pre-training for different models is indicated in Table ~\ref{tab:models_language_stats}.

\iffalse
\begin{itemize}[topsep=2pt,noitemsep]
\item \textbf{Llama3} \cite{dubey2024llama}:\\ 
\textsc{Meta-Llama-3-8B-Instruct}\\
\textsc{Meta-Llama-3-70B-Instruct}\\
\textsc{Meta-Llama-3-70B}
\item \textbf{Mixtral} \cite{jiang2024mixtral}: \\
\textsc{Mixtral-8x7B-Instruct}\\
\textsc{Mixtral-8x22B-Instruct}

\item \textbf{EuroLLM} \cite{martins2024eurollm}: \\ 
\textsc{EuroLLM-1.7B-Instruct}\\
\textsc{EuroLLM-1.7B}%\footnote
\item \textbf{GPT-SW3} \cite{ekgren-etal-2024-gpt}: \\ 
\textsc{GPT-SW3-20b-Instruct}\\
\textsc{GPT-SW3-20b}\\
\textsc{GPT-SW3-40b}
\end{itemize} 
\fi
% 

As our goal is to evaluate the inherent ability of the base model to detect hallucinations, we refrain from model fine-tuning on relevant data and few-shot prompting. After experimentation on the trial sets, the following generation parameters were used for all models: temperature = $0.1$, top-k sampling = $20$, maximum number of generated tokens = $5$. Information about the computational efficiency of our experiments can be found in Appendix ~\ref{app:compute_env}. %All model settings were determined based on results on the trial sets.

\begin{table*}[t]
    \centering
    \begin{small}
    \begin{tabular}{p{1.2cm} p{10.1cm} p{3.5cm}} %
    \toprule
    \textbf{Prompt Name} & \textbf{Prompt} & \textbf{Example output} \\
    \midrule
    Prompt 1 & Given a source sentence (src) and two <scenario> hypotheses (hyp1 and hyp2), detect which of the two is a hallucination of the src. Hallucination means that the hypothesis is not logically supported by the src. &  ``hypothesis1'' $\Rightarrow$ \color{blue}{\verb|hyp1|} \\
    \midrule
    Prompt 2 & You are an AI judge specialised in <scenario> detection. 
    
    Your task is the following: Given a source sentence (src) and two <scenario> hypotheses (hyp1 and hyp2), detect which of the two is a hallucination of the src. Hallucination means that the hypothesis is not logically supported by the src. &  ``The answer is hyp2'' $\Rightarrow$ \color{blue}{\verb|hyp2|} \\
    \midrule
    Prompt 3 & Given a source sentence (src) and two <scenario> hypotheses (hyp1 and hyp2), detect which of the two is a hallucination of the src.  & ``second'' $\Rightarrow$ \color{blue}{\verb|hyp2|} \\
    \midrule
    Prompt 4 & Given a source sentence (src) and two <scenario> hypotheses (hyp1 and hyp2), detect which one of the two logically contradicts the src. & ``both'' $\Rightarrow$ \color{red}{invalid}  \\
    \midrule
    Prompt 5 & Given a source sentence (src) and two <scenario> hypotheses (hyp1 and hyp2), detect which one of the two supports the src. & ``2'' $\Rightarrow$ \verb|hyp2| $\Rightarrow$ \color{blue}{\verb|hyp1|}\color{black}{*}\\
    \midrule
    Prompt 6 & Given a source sentence (src) and two paraphrase hypotheses (hyp1 and hyp2), judge which of the two is a bad <scenario> of the src. & ``Hypothesis'' $\Rightarrow$ \color{red}{invalid} \\
    \midrule
    Prompt 6 & You are an AI judge with expertise in machine translation. Given a source sentence (src) and two translation hypotheses (hyp1 and hyp2), your task is to judge which of the two is a bad translation of the source. & ``It's hard to say'' $\Rightarrow$ \color{red}{invalid}\\
    \bottomrule
    \end{tabular}
    \end{small}
    \caption{Prompt formulations in English tested on all models. For prompts 1-5 <scenario> is replaced with ``paraphrase'' or ``translation''. The last column shows example of generated outputs (translated to English when needed) and the label extracted by post-processing. These examples occur across all prompt variations and are not limited to the prompt they appear next to. *Note that Prompt 5 is a special case where the label is flipped.}
    \label{tab:detection_prompts}
\end{table*}

\begin{table*}[t]
\scriptsize
\centering
%{\renewcommand{\arraystretch}{1.3}%
\resizebox{0.8\textwidth}{!}{%
%\parbox{.45\linewidth}{
%{\renewcommand{\arraystretch}{0.2}%
\begin{tabular}{|l|c|c|c|c|c|c|c|c| } 
\hline
  \rowcolor{gray!30}
  \multicolumn{9}{|l|}{\textbf{English paraphrase}} \\
  \hline
  %MoritzLaurer/
  \textsc{BGE-M3-zeroshot-v2.0} & \multicolumn{1}{c}{\textbf{0.90}} & \multicolumn{7}{r|}{} \\
    \hline
  \rowcolor{gray!30}
  \textbf{LLM} & \textbf{PLg} & \textbf{P1} & \textbf{P2} & \textbf{P3} & \textbf{P4} & \textbf{P5} & \textbf{P6} & \textbf{Avg $\pm$ SD}\\
  \hline
  %meta-llama/
  \textsc{Meta-Llama-3-8B-Instruct} & en & 0.43 & 0.44 & 0.35 & 0.37 & 0.87 & 0.60 & 0.51 $\pm$ 0.20 \\
  \hline
  %meta-llama/
  \textsc{Meta-Llama-3-70B-Instruct} & en & \textbf{0.84} & \textbf{0.92} & 0.69 &	\textbf{0.88} &\textbf{0.94} & \textbf{0.91} & \textbf{0.86} $\pm$ 0.09 \\
  \hline
  %meta-llama/
  \textsc{Meta-Llama-3-70B} & en & 0.70 & 0.58 & 0.59 & 0.70 & 0.63 & 0.81 & 0.67 $\pm$  0.09 \\
  \hline
  %mistralai/
  \textsc{Mixtral-8x7B-Instruct} & en & 0.76 & 0.79 & \textbf{0.81} &0.80 & 
  %0.83
  0.82 &0.86 & 0.81 $\pm$ 0.03\\
  \hline
  %mistralai/
  \textsc{Mixtral-8x22B-Instruct}& en & 0.48 & 0.77 & 0.50 & 0.41 & 
  % 0.80
  0.85 & 0.76 & 0.63 $\pm$ 0.19 \\
  \hline
  %utter-project/
  \textsc{EuroLLM-1.7B-Instruct} & en & 0.32	& 0.41	& 0.28 &	0.33 &	
  % 0.568 -> 0.57 (this didn't change but maybe was copied wrongly without rounding)
  %0.56 
  0.57 & 0.29 & 0.37 $\pm$ 0.11 \\
  \hline
  %utter-project/
  \textsc{EuroLLM-1.7B} & en & 0.45 & 0.45 & 0.46 & 0.45 &  0.22 & 0.45 & 0.41 $\pm$  0.09 \\
  \hline 
  %AI-Sweden-Models/
  \textsc{GPT-SW3-20b-Instruct} & en & 0.45 & 0.07 & 0.45 & 0.44 & 0.22 & 0.44 & 0.35 $\pm$ 0.16 \\ 
  \hline
  %AI-Sweden-Models/
  \textsc{GPT-SW3-20b} & en & 0.55 & 0.44 & 0.48 & 0.50 & 0.31 & 0.52 & 0.47 $\pm$ 0.09 \\
  \hline
  %AI-Sweden-Models/
  \textsc{GPT-SW3-40b} & en & 0.27 & 0.22 & 0.31 & 0.22 & 
  0.50 & 0.23 & 0.29 $\pm$ 0.11 \\
  \hline\hline
  \rowcolor{gray!30}
  \multicolumn{9}{|l|}{\textbf{Swedish paraphrase}} \\
  \hline
  %MoritzLaurer/
  \textsc{BGE-M3-Zeroshot-v2.0} & \multicolumn{1}{c}{\textbf{0.92}} & \multicolumn{7}{c|}{} \\ 
  \hline
  %alexandrainst/
  \textsc{Scandi-NLI-Large} & \multicolumn{1}{c}{\textbf{0.92}} & \multicolumn{7}{c|}{} \\
  \hline 
  \rowcolor{gray!30}
  \textbf{LLM} & \textbf{PLg} & \textbf{P1} & \textbf{P2} & \textbf{P3} & \textbf{P4} & \textbf{P5} & \textbf{P6} & \textbf{Avg $\pm$ SD}\\
  \hline
  \multirow{2}{*}{%meta-llama/
  \textsc{Meta-Llama-3-8B-Instruct}} & en &  0.49 & 0.56 & 0.49 & 0.53 & 0.58 & 0.50 & 0.52 $\pm$ 0.04 \\
                       & sv & 0.40 & 0.47 & 0.45 & 0.42 &  0.69 & 0.49 & 0.49 $\pm$ 0.10 \\
                       \hline
                       
  \multirow{2}{*}{%meta-llama/
  \textsc{Meta-Llama-3-70B-Instruct}} & en & 0.72 & \textbf{0.86} & 0.62 &	0.76 & 0.80 &	0.78 & 0.76 $\pm$ 0.04\\
    & sv & \textbf{0.79} & 0.81 & 0.46 &	0.65 & \textbf{0.83} & 0.83 & 0.73 $\pm$ 0.03\\
  \hline
  \multirow{2}{*}{%meta-llama/
  \textsc{Meta-Llama-3-70B}} & en & 0.54 & 0.45 & 0.55 & 0.63 & 0.56 & 0.63 & 0.56 $\pm$ 0.07 \\
   & sv  & 0.36 & 0.32 & 0.33 & 0.41 & 0.57 & 0.50 & 0.42 $\pm$ 0.10 \\
\hline                              
  \multirow{2}{*}{%mistralai/
  \textsc{Mixtral-8x7B-Instruct }} & en & \textbf{0.79} & 0.84 &	\textbf{0.85} &0.80 &	
  % 0.82
  0.81 &	\textbf{0.86} &\textbf{0.83}  $\pm$ 0.05 \\
  & sv & 0.78 &0.75&	0.74 &	\textbf{0.88} &
  % 0.80
  0.79 &	0.66 & 0.77 $\pm$ 0.08 \\
  \hline
  
\multirow{2}{*}{%mistralai/
\textsc{Mixtral-8x22B-Instruct}} & en & 0.44 & 0.71 & 0.46 & 0.39 & 
% 0.74
0.77 & 0.69 & 0.58 $\pm$ 0.17 \\
 & sv & 0.38 & 0.34 & 0.28 & 0.40 & 
 % 0.80
 0.79 & 0.09 & 0.38 $\pm$ 0.23 \\
  \hline
  
  \multirow{2}{*}{%utter-project/
  \textsc{EuroLLM-1.7B-Instruct}} & en & 0.62	& 0.62 &	0.55	& 0.63 &	0.39 &	0.60 & 0.57 $\pm$ 0.01\\
  & sv & 0.34	& 0.33 &	0.33	& 0.33 &	
  % 0.29
  0.32 &	0.33 & 0.33 $\pm$ 0.01\\
  \hline
  
  \multirow{2}{*}{%utter-project/
  \textsc{EuroLLM-1.7B}} &en & 0.34 & 0.32 & 0.34 & 0.34 & 0.33 & 0.34 & 0.34 $\pm$ 0.00 \\
  & sv &  0.33 & 0.34 & 0.33 & 0.34 & 0.33 & 0.33 & 0.33 $\pm$ 0.00 \\
  \hline

  \multirow{2}{*}{%AI-Sweden-Models/
  \textsc{GPT-SW3-20b-Instruct}} & en & 0.33 & 0.14 & 0.33 & 0.33 & 
  % 0.33
  0.32 & 0.33 & 0.30 $\pm$ 0.08 \\
 & sv & 0.01 & 0.04 & 0.03 & 0.04 & 
 % 0.12
 0.32 & 0.33 & 0.13 $\pm$ 0.15 \\
  \hline
   \multirow{2}{*}{%AI-Sweden-Models/
   \textsc{GPT-SW3-20b}} & en & 0.33 & 0.15 & 0.33 & 0.40 & 0.33 & 0.32 & 0.31 $\pm$ 0.08 \\
 & sv  & 0.39 & 0.33 & 0.37 & 0.35 & 0.32 & 0.36 & 0.35 $\pm$ 0.03 \\
  \hline
  
  \multirow{2}{*}{%AI-Sweden-Models/
  \textsc{GPT-SW3-40b}} & en & 0.43 & 0.34 & 0.5 & 0.41 & 0.45 & 0.52 & 0.44 $\pm$ 0.06 \\
  & sv & 0.45 & 0.39 & 0.53 & 0.50 &  
  % 0.41
  0.41 & 0.40 & 0.45 $\pm$ 0.06 \\
  \hline
 \end{tabular}%
}
\caption{Test set results for the paraphrase scenario in English and Swedish: F1 scores. Baseline models have a single score. For all other models, we report scores for different combinations of prompt language (PLg) and prompt formulation (P1--P6), as well as (Avg) and standard deviation (SD). Boldface marks highest score per column.}
\label{tab:detection_paraphrase}
\end{table*}

\subsection{Prompting}
\label{subsec:experimental_variables}
To investigate how model performance depends on the specific formulation of the prompt, we experiment with six different prompting strategies, exemplified in Table~\ref{tab:detection_prompts}. 
The prompts differ with respect to whether they explicitly mention the term ``hallucination'' (Prompts 1--3 vs.\ 4--6) and whether they include an explicit definition of the concept of hallucination (Prompts 1--2 vs.\ 3--6). Prompts 4--6 (which contain neither the term ``hallucination'' nor an explicit definition) use formulations that to different degrees approximate the notion of hallucination with terms like ``contradicts'', ``supports'' and ``bad''. Note that the formulation with ``support'' inverts the task by prompting the model to identify the good hypothesis rather than the hallucination, which needs to be handled in post-processing to make sure that the evaluation is correct (see Appendix~\ref{sec:post-processing}). An additional variable is the language of the prompt: we experiment with prompting in English versus %the language of the example (or source sentence for paraphrase). %the language present in the examples. %desired output language. 
the language of the source sentence (which in the case of paraphrase is also the target language).
%Table~\ref{tab:detection_prompts} presents the complete list of prompt formulations tested in English. %All prompt formulations are also tested in Swedish, French, and German, depending on the language of the source sentence. 
Prompts in Swedish, French, and German can be found in Table~\ref{tab:non_English_prompts} in Appendix ~\ref{sec:app_prompts}. 

In addition to the base prompts, all models receive a near identical set of instructions to provide only ``hyp1'' or ``hyp2'' as acceptable answers and to start the text generation with ``The answer is:'' (or a similar phrase). Differences in the additional prompt instructions are minimal; they vary only by language or phrasing  depending on the model. Though we did not prompt the models to do so, they sometimes provide explanations of the output.%No explanation was requested of the models although they sometimes generate explanations for the output even when not prompted to do so. 

\subsection{Evaluation}
All models are evaluated with respect to the gold labels in the datasets, using 
the F1 metric.
% the following metrics: accuracy, precision, recall, and F1 score (the primary metric). 
The model output first undergoes simple rule-based post-processing to check for produced labels in a number of variations and map them to $hyp1$ or $hyp2$ (e.g. ``hypothesis 1'' or ``första'' for hypothesis 1, and ``hypothesis 2'' or ``zweite'' for hypothesis 2).
Model outputs are considered invalid in cases where the model produces either no label at all or a label outside of the allowed set: \{$hyp1$, $hyp2$\}. Examples of outputs produced during the experiments can be found in Table \ref{tab:detection_prompts}.
% \footnote{The post-processing also handles the special case of Prompt 5, noted earlier, where predictions need to flipped.} 
The post-processing is described in more detail in Appendix~\ref{sec:post-processing}. 

\subsection{NLI Baseline}

As baselines, we use NLI models, which are computationally inexpensive and trained specifically for predicting textual entailment. % They have also been shown to have promising performance in hallucination detection \citep{manakul2023selfcheckgpt,Kang2024ComparingHD,mickus2024semeval}. 
NLI models typically classify a sentence pair into one of three classes: entailment, neutral, and contradiction. We selected two multilingual zero-shot NLI models with no ``neutral'' label, meaning they only predict the textual entailment between a premise and a hypothesis. The baseline used for all scenarios is \textsc{BGE-M3-Zeroshot-v2.0}, a multilingual zero-shot XLM-RoBERTa model
% that builds on
based on BGE M3-Embeddings \cite{bge-m3}. An additional NLI baseline for the Swedish paraphrase scenario is \textsc{Scandi-NLI-Large} \cite{scandi-nli}, which is trained on Swedish, Danish, and Norwegian data. 
%\verb|MoritzLaurer/bge-m3-zeroshot-v2.0| 
We first predict  ``entailment`` and ``not\_entailment'' class scores between the source sentence and each hypothesis. We infer the label based on the predicted entailment value for each of the two hypotheses. More details can be found in Section~\ref{sec:nli_appendix} in the Appendix.

\begin{comment}
To determine which of the two hypotheses ($hyp1$, $hyp2$) contains a hallucination, we predict  ``entailment`` (E) and ``not\_entailment'' (NE) class scores between the source sentence and each one of the hypotheses. We then apply the following logic to choose the hallucination:
We then choose the hallucination based on which one or more hypotheses 
\begin{itemize}[topsep=2pt,noitemsep]
    \item If \textbf{E} $>$ \textbf{NE} for one hypothesis and \textbf{E} $<$ \textbf{NE} for the other, we choose the one with \textbf{E} $<$ \textbf{NE}.
    \item If \textbf{E} $>$ \textbf{NE} for both hypotheses, we choose the one with the lowest \textbf{E} score.
    \item If \textbf{E} $<$ \textbf{NE} for both hypotheses, we choose the one with the highest \textbf{NE} score.
\end{itemize}

\[
\text{hypothesis with}
\parbox{\columnwidth}{$
    \begin{cases} 
        E < NE, & \text{if one has } E > NE \text{the other } E < NE, \\
        \text{lowest } E, & \text{if both have } E > NE, \\
        \text{highest } NE, & \text{if both have } E < NE.
    \end{cases}
$}
\]
\end{comment}

The default configurations are used for both models and each pair ($source$+$hyp1$ / $source$+$hyp2$). For the translations, the \textsc{BGE-M3-Zeroshot-v2.0} NLI model receives two sentences in two different languages as input (one in English, and one in French or German) in both directions.

\begin{table*}[!ht]
\scriptsize
\centering
%{\renewcommand{\arraystretch}{1.3}%
\resizebox{0.8\textwidth}{!}{%
%\parbox{.45\linewidth}{
%{\renewcommand{\arraystretch}{0.2}%
\begin{tabular}{|l|c|c|c|c|c|c|c|c|} 
\hline
  \rowcolor{gray!30}
  \multicolumn{9}{|l|}{\textbf{Translation en$\Rightarrow$fr}} \\
  \hline
  %MoritzLaurer/
  \textsc{BGE-M3-Zeroshot-v2.0} & \multicolumn{1}{c}{0.82} & \multicolumn{7}{r|}{} \\
    \hline
  \rowcolor{gray!30}
  \textbf{LLM} & \textbf{PLg} & \textbf{P1} & \textbf{P2} & \textbf{P3} & \textbf{P4} & \textbf{P5} & \textbf{P6} & \textbf{Avg $\pm$ SD}\\
  \hline
  %meta-llama/
  \textsc{Meta-Llama-3-8B-Instruct} & en & 0.74 & 0.77 & 0.66 & 0.71 & 0.83 & 0.73 & 0.74 $\pm$ 0.06 \\
  \hline
  %meta-llama/
  \textsc{Meta-Llama-3-70B-Instruct} & en & \textbf{0.85}	&
  \textbf{0.89} & 0.81 & \textbf{0.88} & \textbf{0.86} & \textbf{0.90} & \textbf{0.87} $\pm$ 0.03 \\
  \hline
  %meta-llama/
  \textsc{Meta-Llama-3-70B} & en & 0.69 & 0.73 & 0.70 & 0.74 & 0.49 & 0.74 & 0.68 $\pm$ 0.10 \\
  \hline
  %mistralai/
  \textsc{Mixtral-8x7B-Instruct} & en & 0.81	& 0.86 & \textbf{0.85} &	0.78 &	
  % 0.83 0.82
  0.83 &	0.80 & 0.82 $\pm$ 0.03 \\
   \hline
   %mistralai/
   \textsc{Mixtral-8x22B-Instruct} & en & 0.41 & 0.68 & 0.57 & 0.44 & 
   % 0.74 0.78
   0.74 & 0.45 & 0.56 $\pm$ 0.15 \\
  \hline
  %utter-project/
  \textsc{EuroLLM-1.7B-Instruct} & en & 0.34	& 0.44 &0.49 &0.40& 0.60	& 0.49 & 0.46 $\pm$ 0.09 \\
   \hline
   %utter-project/
   \textsc{EuroLLM-1.7B} & en & 0.44 & 0.42 & 0.44 & 0.43 &
   % 0.23 (it actually used to be 0.24, maybe a bad copy without rounding?)
   0.23 & 0.43 & 0.40 $\pm$ 0.08 \\
    \hline\hline
  \rowcolor{gray!30}
  \multicolumn{9}{|l|}{\textbf{Translation fr$\Rightarrow$en}} \\
  \hline
  %MoritzLaurer/
  \textsc{BGE-M3-Zeroshot-v2.0} & \multicolumn{1}{c}{\textbf{0.88}} & \multicolumn{7}{r|}{} \\
    \hline
  \rowcolor{gray!30}
  \textbf{LLM} & \textbf{PLg} & \textbf{P1} & \textbf{P2} & \textbf{P3} & \textbf{P4} & \textbf{P5} & \textbf{P6} & \textbf{Avg $\pm$ SD}\\
  \hline
  \multirow{2}{*}{%meta-llama/
  \textsc{Meta-Llama-3-8B-Instruct}} & en & 0.62 & 0.63 & 0.53 & 0.60 & 0.73 & 0.57 & 0.61 $\pm$ 0.07 \\
  & fr & 0.33 & 0.40 & 0.30 & 0.43 &  0.80 & 0.73 & 0.50 $\pm$ 0.21 \\
  \hline
  \multirow{2}{*}{%meta-llama/
  \textsc{Meta-Llama-3-70B-Instruct}} & en & 0.67 & 0.80 & 0.53 &\textbf{0.84} & \textbf{0.81} &	0.78 & 0.74 $\pm$ 0.12\\
    & fr & 0.80 & 0.80 & 0.73 &	\textbf{0.84} & \textbf{0.81} &	0.80 & \textbf{0.80} $\pm$ 0.04\\
  \hline
  \multirow{2}{*}{%meta-llama/
  \textsc{Meta-Llama-3-70B}} & en &
  0.63 & 0.70 & 0.58 & 0.68 & 0.61 & 0.66 & 0.64 $\pm$ 0.05 \\
 & fr  & 0.50 & 0.62 & 0.41 & 0.41 & 0.51 & 0.75 & 0.53 $\pm$ 0.13 \\
\hline

  \multirow{2}{*}{%mistralai/
  \textsc{Mixtral-8x7B-Instruct }} & en & 0.80	& \textbf{0.82}	& 0.78 &	0.83 &	
  % 0.81 0.77
  \textbf{0.81} &	\textbf{0.81} & \textbf{0.80} $\pm$ 0.02 \\
  & fr & \textbf{0.81}	&0.77&	\textbf{0.85}	&0.78&
  % 0.80 0.79
  0.80 &	0.78 & \textbf{0.80} $\pm$ 0.03 \\
  \hline

  \multirow{2}{*}{%mistralai/
  \textsc{Mixtral-8x22B-Instruct}} & en & 0.39 & 0.56 & 0.46 & 0.56 & 
  % 0.72 0.80
  0.72 & 0.41 & 0.53 $\pm$ 0.15 \\
  & fr &  0.07 & 0.26 & 0.05 & 0.13 & 
  % 0.53 0.59
  0.53 & 0.34 & 0.24 $\pm$ 0.20 \\
  \hline
  
  \multirow{2}{*}{%utter-project/
  \textsc{EuroLLM-1.7B-Instruct}} & en & 0.40	& 0.52 &	0.46 & 0.40 & 0.38	& 0.51 & 0.45 $\pm$ 0.06\\
  & fr & 0.35	&0.36	&0.32&	0.34&	
  % 0.31 0.33
  0.31	&0.35& 0.34 $\pm$ 0.01\\
  \hline

  \multirow{2}{*}{%utter-project/
  \textsc{EuroLLM-1.7B}} & en & 0.35 & 0.35 & 0.35 & 0.36 & 0.31 & 0.35 & 0.35 $\pm$ 0.02 \\
  & fr &  0.35 & 0.34 & 0.35 & 0.34 & 
  %  0.31
  0.31 & 0.34 & 0.34 $\pm$ 0.01 \\
  \hline
  \hline
  \rowcolor{gray!30}
  \multicolumn{9}{|l|}{\textbf{Translation en$\Rightarrow$de}} \\
  \hline
  %MoritzLaurer/
  \textsc{BGE-M3-Zeroshot-v2.0} & \multicolumn{1}{c}{0.73} & \multicolumn{7}{c|}{} \\
  \hline 
    \rowcolor{gray!30}
  \textbf{LLM} & \textbf{PLg} & \textbf{P1} & \textbf{P2} & \textbf{P3} & \textbf{P4} & \textbf{P5} & \textbf{P6} & \textbf{Avg $\pm$ SD}\\
  \hline
   %meta-llama/
   \textsc{Meta-Llama-3-8B-Instruct} & en & 0.56 & 0.62 & 0.48 & 0.57 & 0.79 & 0.60 & 0.60 $\pm$ 0.10 \\
   \hline
   %meta-llama/
   \textsc{Meta-Llama-3-70B-Instruct} & en & 0.69	& \textbf{0.87}	&0.68	&\textbf{0.75}	& 0.83	& \textbf{0.85} & 0.78 $\pm$ 0.08 \\
   \hline
   %meta-llama/
   \textsc{Meta-Llama-3-70B} & en & 0.65 & 0.70 & 0.61 & 0.65 & 0.54 & 0.81 & 0.66 $\pm$ 0.09 \\
   \hline
   %mistralai/
   \textsc{Mixtral-8x7B-Instruct} & en & \textbf{0.82} &	0.79 &	\textbf{0.78}& \textbf{0.75} &	
   % 0.84 0.83
   0.84 & 0.79 & \textbf{0.79} $\pm$ 0.03 \\
   \hline
   %mistralai/
   \textsc{Mixtral-8x22B-Instruct} & en & 0.49 & 0.75 & 0.64 & 0.57 & 
   % 0.81 0.87
   \textbf{0.81} & 0.59 & 0.65 $\pm$ 0.14 \\
   \hline 
   %utter-project/
   \textsc{EuroLLM-1.7B-Instruct} & en & 0.33&	0.45&	0.40	&0.41	&0.53&	0.46 & 0.43 $\pm$ 0.07 \\
  \hline
  %utter-project/
  \textsc{EuroLLM-1.7B} & en & 0.42 & 0.41 & 0.42 & 0.42 & 0.24 & 0.42 & 0.39 $\pm$ 0.07 \\
  \hline\hline
    \rowcolor{gray!30}
  \multicolumn{9}{|l|}{\textbf{Translation de$\Rightarrow$en}} \\
  \hline
  %MoritzLaurer/
  \textsc{BGE-M3-Zeroshot-v2.0} & \multicolumn{1}{c}{0.78} & \multicolumn{7}{r|}{} \\
    \hline
  \rowcolor{gray!30}
  \textbf{LLM} & \textbf{PLg} & \textbf{P1} & \textbf{P2} & \textbf{P3} & \textbf{P4} & \textbf{P5} & \textbf{P6} & \textbf{Avg $\pm$ SD}\\
  \hline
  \multirow{2}{*}{%meta-llama/
  \textsc{Meta-Llama-3-8B-Instruct}} & en &0.56& 0.58& 0.46& 0.52&  0.79& 0.47& 0.57 $\pm$ 0.12\\
  & de & 0.41& 0.36& 0.19& 0.48&  0.80& 0.67& 0.49 $\pm$ 0.22 \\
\hline
   \multirow{2}{*}{%meta-llama/
   \textsc{Meta-Llama-3-70B-Instruct}} & en & 0.66& 0.85& 0.60&	0.82& 0.81&	\textbf{0.85}& 0.77 $\pm$ 0.11\\
    & de & 0.53& \textbf{0.87}& 0.20& \textbf{0.86}& \textbf{0.83}&	0.83& 0.69 $\pm$ 0.27\\
  \hline
  \multirow{2}{*}{%meta-llama/
  \textsc{Meta-Llama-3-70B}} & en & 0.56& 0.57& 0.50& 0.55& 0.67& 0.60&0.58 $\pm$ 0.06 \\
 & de & 0.34& 0.72& 0.30& 0.38& 0.67& 0.56& 0.49 $\pm$ 0.18 \\
 \hline
  \multirow{2}{*}{%mistralai/
  \textsc{Mixtral-8x7B-Instruct }} & en & 0.75&0.82&\textbf{0.85}&0.85&	
  %  0.81 0.82
  0.81 &	0.84& \textbf{0.82} $\pm$ 0.04 \\
  & de & \textbf{0.81}& 0.80&	0.81&0.77&
  %   0.84 0.83
  0.84 &	0.62& 0.77 $\pm$ 0.08 \\
  \hline

  \multirow{2}{*}{%mistralai/
  \textsc{Mixtral-8x22B-Instruct }} & en & 0.43& 0.58& 0.42& 0.56&
  % 0.79 0.86
  0.79 & 0.37& 0.54 $\pm$ 0.18 \\
  & de &  0.18& 0.38& 0.33& 0.19& 
  % 0.76 0.81
  0.76 & 0.57& 0.41 $\pm$ 0.24 \\
  \hline
  
  \multirow{2}{*}{%utter-project/
  \textsc{EuroLLM-1.7B-Instruct}} & en & 0.22&0.23&	0.21&0.22&0.46&0.21& 0.26 $\pm$ 0.10\\
  & de & 0.20&0.22&0.22&0.22&
  % 0.45 0.46
  0.45 &0.22& 0.26 $\pm$ 0.10\\
  \hline
  \multirow{2}{*}{%utter-project/
  \textsc{EuroLLM-1.7B}} & en & 0.45& 0.39& 0.41& 0.48& 0.30& 0.47& 0.42 $\pm$ 0.07 \\
  & de & 0.24& 0.22& 0.28& 0.25& 
  % 0.46
  0.46& 0.22& 0.28 $\pm$ 0.09 \\
\hline
\end{tabular}%
}
\caption{Test set results for the translation scenario in all language pairs: F1 scores. Baseline models have a single score. For all other models, we report scores for different combinations of prompt language (PLg) and prompt formulation (P1--P6), as well as (Avg) and standard deviation (SD). Boldface marks highest score per column.
}
\label{tab:detection_translation}
\end{table*}

\section{Results}
\label{sec:results}

Tables~\ref{tab:detection_paraphrase} and~\ref{tab:detection_translation} present model scores for different prompt formulations and prompt languages in the paraphrase and translation scenarios. Overall, we observe that performance varies considerably between models. We also note that the NLI baseline is %competitive with, if not better than, the LLMs on this task. 
hard to beat, especially in the paraphrase scenario and for translation from French to English. This corroborates the findings of %the HalluciGen task 
\citet{durlich2024overview}. % The results for each of the two scenarios are discussed below.

\subsection{Paraphrase}

For English paraphrases, we observe that \textsc{Meta-Llama-3-70B-Instruct} has the strongest overall performance, although with three of the prompts it does not beat the NLI baseline. The competitive performance of the NLI baseline is even more apparent in the Swedish paraphrase scenario, where 
%even 
the best-performing LLMs (\textsc{Meta-Llama-3-70B-Instruct} and \textsc{Mixtral-8x7B-Instruct}) are outperformed by the NLI baseline, irrespective of the prompt used. All GPT-SW3 models perform poorly for both Swedish and English. A striking observation is that the performance of \textsc{GPT-SW3-20b-Instruct} reaches the low F1 score of 0.07 for Prompt 2 for Swedish. When prompted with ``You are an AI judge specialised in \ldots'', \textsc{GPT-SW3-20b-Instruct} provides mostly invalid answers. \textsc{EuroLLM-1.7B-Instruct} exhibits comparable performance with the GPT-SW3 models on English paraphrase, and even surpasses them on Swedish paraphrase. The latter is surprising given %the larger model size and 
the larger amount of Swedish data in the GPT-SW3 models. %in comparison to \textsc{EuroLLM-1.7B-Instruct}. 
Lastly, the performance of \textsc{EuroLLM-1.7B} is generally on par with \textsc{GPT-SW3-20b}.  

%or English paraphrase, the variance in the performance of \textsc{Meta-Llama-3-8B-Instruct}, \textsc{Mixtral-8x22B-Instruct}, \textsc{EuroLLM-1.7B-Instruct}, and \textsc{gpt-sw3-20b-instruct} is quite high. Prompt 5 leads to the best performance for 

%Considering the standard deviation values in the \textit{avg score} column in Table~\ref{tab:detection_paraphrase}, for some models, like \textsc{Mixtral-8x7B-Instruct}, \textsc{Meta-Llama-3-70B-Instruct} and \textsc{Meta-Llama-3-70B}, the performance is quite stable over the 6 different prompts, while for \textsc{Meta-Llama-3-8B-Instruct}, \textsc{Mixtral-8x22B-Instruct}, \textsc{EuroLLM-1.7B-Instruct}, and \textsc{gpt-sw3-20b-instruct} (English) the variance is quite high.
%Furthermore, the performance of a given LLM does not vary significantly between prompts, except \textsc{Meta-Llama-3-8B-Instruct}, \textsc{gpt-sw3-20b-instruct} and \textsc{EuroLLM-1.7B-Instruct}, indicated by the high standard deviation values in the \textit{avg score} column in Table~\ref{tab:detection_paraphrase}.The performance of those four models stands out for Prompt 5 -- whether the hypothesis supports the source, with no mention of the term hallucination: this prompt works best for \textsc{Meta-Llama-3-8B-Instruct}, \textsc{Mixtral-8x22B-Instruct} and \textsc{EuroLLM-1.7B-Instruct} in English and for the first two even with Swedish paraphrases, but it leads to low scores for \textsc{gpt-sw3-20b-instruct}. 

\subsection{Machine Translation}

In the Machine Translation scenario, %for which we have a smaller set of models, 
we again observe stronger performance for \textsc{Mixtral-8x7B-Instruct} and \textsc{Meta-Llama-3-70B-Instruct} compared with \textsc{EuroLLM-1.7B-Instruct}.
In contrast with the paraphrase scenario, where we observe that the NLI baseline often outperforms even the strongest LLMs, for translation we almost see the opposite: the NLI baseline is outperformed by either \textsc{Meta-Llama-3-70B-Instruct} or \textsc{Mixtral-8x7B-Instruct} for every language direction except fr$\Rightarrow$en. One obvious difference is that whilst the paraphrase task is monolingual, the cross-lingual nature of the translation task adds complexity, as the model not only needs to perform the NLI task but also implicit translation. As translation examples are likely present in pre-training data, and possibly addressed by subsequent instruction-tuning, %is very likely one of the tasks that multilingual LLMs are specifically trained for (in addition to NLI),
this may give LLMs an edge over NLI models. Further investigation is needed to determine whether this is the case.

\section{Discussion}
\label{sec:discussion}
The results presented in Section~\ref{sec:results} support the use of LLMs, and also NLI models, for the hallucination detection task. We now discuss the %results %with respect to our research questions 
%regarding 
differences in performance across target languages as well as the effects of model size, instruction tuning, and the language and formulation of the prompts.

%\paragraph{How does model performance differ across target languages?}
\subsection{Research Questions}
\paragraph{How does model performance on hallucination detection differ between target languages?}

%To address this question we look for high-level trends that are common across the different target languages. 
We find that the capability of the model to detect hallucinations is generally consistent between target languages, with often a slight performance benefit 
%when the source sentences are in English. 
for English source sentences.
This is not surprising given that English is most likely the dominant language in the data used for pre-training and instruction tuning of the models. %under study. 
Two exceptions are \textsc{GPT-SW3-40b} and \textsc{EuroLLM-1.7B-Instruct} that both have better performance on Swedish than English, despite being trained on larger amounts of English data compared to Swedish. %in comparison to Swedish. 
In addition,  
%A direct performance comparison between models in relation to the amount of target language data present in their pre-training is only possible for the EuroLLM and GPT-SW3 models. That is due to the lack of available information about the pre-training data distribution of Llama3 and Mixtral models. 
it is observed that \textsc{EuroLLM-1.7B-Instruct} outperforms all three GPT-SW3 models on the Swedish paraphrase scenario, despite the limited amount of Swedish pre-training data in the former model. This indicates that the amount of target language data used in pre-training is not the sole factor contributing to the model performance on hallucination detection in languages other than English.

\paragraph{Does increased model parameter size lead to better performance?}

To address this question, we compare the performance of models with different numbers of parameters belonging to the same family. For Llama3 we observe that model size has a clear impact, with the larger \textsc{Meta-Llama-3-70B-Instruct} model outperforming the smaller \textsc{Meta-Llama-3-8B-Instruct} model, typically by a large margin. We see the same pattern for GPT-SW3, but only for Swedish,
%paraphrase, 
where \textsc{GPT-SW3-40b} consistently outperforms the smaller \textsc{GPT-SW3-20b} model. The opposite trend is observed for the Mixtral models: increasing the model size from 8x7b to 8x22b consistently worsens performance across all scenarios. 
%This could be related to the Mixtral model architecture, but together with the mixed results for GPT-SW3, these results suggest that other factors besides model size play a role, which warrants further investigation.

\paragraph{Does instruction tuning lead to better performance?}

%To analyse the effect of instruction tuning on model performance, we compare the instruction-tuned models with their base model counterparts of equal size. 
In the case of the Llama3 family, we observe a clear performance improvement in using the instruction-tuned variant over the base \textsc{Meta-Llama-3-70B} in both scenarios and for all languages. %We observe the opposite effect in the case of GPT-SW3, 
The opposite is observed for GPT-SW3, with \textsc{GPT-SW3-20b} consistently outperforming the instruction-tuned variant on both paraphrase scenarios. % (though the difference is marginal for some prompts).
%One factor that might play a role here 
This could be due to the absence of NLI examples in the instruction-tuning corpus used for training \textsc{GPT-SW3-20b-Instruct} \cite{ekgren-etal-2024-gpt}. % Given the similarity of the NLI task with our hallucination detection task, as also indicated by the strong NLI baseline, we would expect a performance increase after training GPT-SW3 on NLI instruction data.
%Lastly, the results for the \textsc{EuroLLM-1.7B} models are mixed. 
The instruction-tuned variant of \textsc{EuroLLM-1.7B} performs better for Swedish paraphrase and fr$\Rightarrow$en translation, while the reverse is true for English paraphrase and de$\Rightarrow$en translation. This may be attributed to the model's limited capacity, which restricts its ability to fully integrate the instruction tuning data.
%The model's capability to perform the hallucination detection task may be due to its size. 
%We propose that the larger versions of the EuroLLM model, once available, be included in future evaluations.
Overall, we do not find conclusive evidence that instruction tuning improves performance, as the results differ between model families, trained on different instruction tuning datasets. 
% Future experiments with models trained on the same instruction tuning data are needed to disentangle these factors. 

%Greater transparency regarding the \textit{nature} of the instruction tuning of models is required in order to provide further insight.

%Liane: The following paragraph contains the original text that we wrote before we exanded the set of mode
%To analyse the effect of instruction tuning on model performance we compare the performance of \textsc{GPT-SW3-20b} and \textsc{GPT-SW3-20b-Instruct} across different prompts. \textsc{GPT-SW3-20b} either marginally outperforms or exhibits the same performance as \textsc{GPT-SW3-20b-Instruct} on the English paraphrase task. The same observation holds for Swedish paraphrase, with the exception of Prompt 2, where \textsc{GPT-SW3-20b-Instruct} outperforms \textsc{GPT-SW3-20b}. One factor that might play a role here is the absence of NLI examples in the instruction-tuning corpus used for training \textsc{GPT-SW3-20b-Instruct} \cite{ekgren-etal-2024-gpt}. Given the similarity of the NLI task with our hallucination detection task, as also indicated by the strong NLI baseline, we would expect a performance increase after training GPT-SW3 on NLI instruction data. 

\paragraph{Does the language and formulation of the
prompt matter?}

We investigate the effect of non-English prompts for Swedish paraphrase and fr$\Rightarrow$en and de$\Rightarrow$en translation. As indicated by the difference in average model performance between prompt languages in Tables~\ref{tab:detection_paraphrase} and~\ref{tab:detection_translation}, the choice of prompt language matters, with English being overall the best-performing %We also observe that English is the best-performing %preferable 
prompt language. This is not surprising given that all models under study have likely been trained on large amounts of English.  %during both pre-training and instruction tuning. 
%A notable difference is observed between instruct and non-instruct variants of \textsc{GPT-SW3-20b} and \textsc{Meta-Llama-3-70B} models. 
One exception is Swedish paraphrase, where \textsc{GPT-SW3-20b-Instruct} performs best with Swedish prompts. %unlike its base variant. 
The same holds for \textsc{Meta-Llama-3-70B-Instruct}, which performs best when prompted in French for fr$\Rightarrow$en translation. %, in contrast to \textsc{Meta-Llama-3-70B}. 

%Considering the best-performing language for each model and scenario, we analyse the effect of varying the prompt formulation according to available choices presented in Table~\ref{tab:detection_prompts}. 
We now investigate whether individual model performance varies with the prompt choice, considering the standard deviation values % in the \textit{Avg $\pm$ SD} column 
in Tables~\ref{tab:detection_paraphrase} and~\ref{tab:detection_translation}. 
%Overall, model performance is stable across varying prompt formulations for all scenarios, but some cases stand out. 
Overall, performance remains stable across prompt variations, but certain cases stand out: \textsc{Mixtral-8x22B-Instruct} is significantly unstable across all scenarios, with Prompt 5 (no mention of ``hallucination'' and use of ``supports'' instead of ``contradicts'') %consistently leading to the best performance. 
consistently performing best. The same
%observation
partially holds for \textsc{Meta-Llama-3-8B-Instruct}. %Finally, we investigate whether the model performance on hallucination detection depends on the presence of the term ``hallucination'' in the prompt, by comparing the overall performance of Prompts 1-–3, which mention ``hallucination'', with Prompts 4--6, which do not. Again, we observe that there is such an effect for \textsc{mistralai/Mixtral-8x22B-Instruct} and some Meta-Llama3 models, as the inclusion of the term ``hallucination'' in the prompt negatively impacts model performance.
Additionally, prompts mentioning ``hallucination'' (Prompts 1–3) tend to negatively impact performance for \textsc{Mixtral-8x22B-Instruct} and some Meta-Llama3 models compared to those that omit it (Prompts 4–6).

\subsection{Error Analysis}

We examine the error rate of each model for different hallucination categories as well as highlighting the proportion of errors caused by the models producing incorrect labels. These results are averaged across all prompts and are shown in detail in Appendix \ref{sec:hallucination-type}.

%Figures \ref{fig:hallucination-type-par} and \ref{fig:hallucination-type-trans} in Appendix \ref{sec:hallucination-type}. In those figures, the hatch represents the proportion of invalid labels. 

% We find that models in general exhibit a higher error rate for natural and pronoun hallucinations in paraphrasing. 

% In general, models tend to exh higher error rates on natural hallucinations and pronouns in the paraphrase scenario (especially the quite large \textsc{Mixtral-8x22B-instruct}).

% The Llama3 models, which exhibit relatively high performance compared to others across both scenarios, show weakness in detecting addition and natural hallucinations (paraphrasing) and gender, conversion, and addition (translation). The larger Mixtral model also shows fluctuations in error rates across hallucination categories, although that could be a result of the overall poor performance caused mostly by invalid output (indicated by the hatches in Figures \ref{fig:hallucination-type-par} and \ref{fig:hallucination-type-trans}). In contrast, GPT-SW3 and EuroLLM models have relatively stable behavior across the different categories.

%However, the error rate can differ largely across hallucination types and fining a distinctive trend for other models seems difficult.
%, both in paraphrasing and translation. 
%This is mostly due to selecting an invalid label (namely by specifying that ``Neither'' hypothesis contained a hallucination). However, its performance continues in a similarly poor trend across different hallucination types. 
We find that the error rate seems to fluctuate across different hallucination categories, but without any strong or discernible patterns. We also find that a  high error rate may be a result of the the number of invalid outputs (i.e., not $hyp1$ nor $hyp2$, nor any synonyms that correspond to either label) produced by some model. We notice this largely in \textsc{Mixtral-8x22B-instruct}, but to a lesser degree in \textsc{gpt-sw3-20b-instruct}, \textsc{Mixtral-8x7B}, and the two fairly small EuroLLM variants (respectively).

%The contrast in performance between \textsc{gpt-sw3-20b} and (to a slight degree) \textsc{EuroLLM-1.7B} against their instruction-tuned variants indicates that instruction-tuning may result in poorer performance for this type of classification tasks.
%Here, the proportion of invalid answers seems consistent across categories. 

Notably, the Mixtral family tends to generate output claiming that both or neither hypotheses are hallucinations. Similarly, GPT\--SW3 models display a habit of returning a near-identical phrase or label for every instance. For example, \textsc{gpt-sw3-20b} tends to detect the $hyp1$ label for nearly every sentence pair, whereas the instruction-tuned variant has a higher error rate caused by invalid outputs, as it tends to, for some prompts, almost always output a phrase indicating its inability to perform the task (e.g.,\textit{``It is hard to say without more context''}\footnote{Originally in Swedish: ``Det är svårt att säga utan mer sammanhang''}). It is unclear why this model tends to converge on near-identical outputs, though it could relate to the type of data used during instruction tuning. Invalid outputs from the EuroLLM models, on the other hand, occur when the models start generating sentences that translate or paraphrase the source sentence instead of performing the detection task at hand, although that is not surprising given their small size. 

It is worth noting that the NLI models' labels are determined by the entailment probability, hence incapable of producing invalid labels within their setup, unlike LLMs.
% In both the paraphrase and machine translation scenarios, we find that a large portion of errors can be attributed to model outputs that are invalid, despite our efforts to post-process the output generously. 

\section{Conclusion}

We have presented a suite of experiments to investigate the capabilities of open-access LLMs for %the task of 
detecting hallucinations, as defined in %our previous work on 
the HalluciGen task \cite{karlgren-eloquent-2024, durlich2024overview}. %We also release the gold labels for this task. 
%Even though most models exhibit varying performance across languages and tasks, 
The strongest models, \textsc{Mixtral-8x7B-Instruct} and \textsc{Meta-Llama-3-70B-Instruct}, perform consistently well across all languages and scenarios, suggesting that LLMs are appropriate for %the hallucination detection 
this task. %We also note the strong performance of the considerably smaller NLI models, especially in the paraphrase scenario, suggesting that LLM-based detectors are not the only viable option. 
The strong performance of the considerably smaller NLI models suggests that LLM-based detectors are not the only viable option. 

We analyse the effect of four different factors: target language, model size, instruction-tuning and prompt -- and find that none of them can be used as a straightforward predictor of model performance on this task. Our controlled experiments indicate that:
\begin{inlinelisttwo}
    \item models perform consistently across languages, with a slight advantage for English;
    \item the impact of model size differs between model families;
    \item instruction-tuning has a clear positive effect only for the largest model% in this comparison
    ;
    \item English prompts generally lead to the best overall performance, while the inclusion of the term ``hallucination'' in the prompt has a partially negative impact; and
    \item for some models, a high error rate can be traced to the proportion of invalid outputs produced during inference.
\end{inlinelisttwo}

In future work, we aim to explore whether LLMs may be used to \textit{generate} datasets for training and evaluating hallucination detectors and apply these in a cross-model evaluation setting. %Furthermore, we may dedicate future work to investigating the performance of LLMs for this task across the different hallucination categories (which we constructed in the dataset). 
% We will also investigate LLM performance across different hallucination categories from our dataset. 
% Finally, we acknowledge the need to further investigate some factors, such as model size, instruction-tuning and pre-training dataset composition by varying one factor consistently across different models. 
%and the composition of the pre-training datasets in a more controlled setup, where we consistently vary the same factor across a wide range of models.  % by evaluating the performance at different training or instruction-tuning checkpoints. 

\begin{comment}
\section*{Acknowledgments}
This work has been partially supported by the Swedish Research Council (grant number 2022-02909) and by UK Research and Innovation (UKRI) under the UK government’s Horizon Europe funding guarantee [grant number 10039436 (Utter)].
\end{comment}

\section*{Limitations}

Owing to the very large and constantly expanding set of available LLMs and the numerous ways in which to prompt them, it is infeasible to conduct exhaustive prompt exploration experiments. In a similar vein, it is infeasible to explore all possible values for the generation parameters described in Section~\ref{subsec:experimental_variables}; though we selected values that should be broadly suitable, we did not optimise these for individual models. Nevertheless, we hope that our work provides insights into the suitability of LLMs as hallucination detectors, as indicated by their performance on the hallucination detection task.

When commenting on the presence of target languages in model pre-training data or the tasks included in instruction-tuning, we are reliant on information provided by the model developers in the form of academic papers, reports, and blog posts. Whilst these aspects are well documented for the EuroLLM and GPT-SW3 models, in the case of other models (e.g. Llama3 and Mixtral) this information may be incomplete or missing. Where such information is not provided, it is naturally difficult to draw conclusions as to the effects of different factors on model performance for any downstream task.

Additionally, two main limitations exist for the hallucination categories labels: (a) they suffer from class imbalance; and (b) they do not take into account that some samples could fall into multiple categories.
% disagreements between annotators were through discussion. 

Our datasets focus only on a small set of high-resourced languages: English and Swedish for paraphrase and the English-French and English-German pairs for translation. Furthermore, a number of hallucination examples were constructed manually and may not accurately reflect real-world intrinsic hallucinations. Future work should look to reduce the English-centric nature of the datasets and expand the task to include a range of high, medium, and low-resource languages with exclusive focus on naturally occurring intrinsic hallucinations.

\bibliography{eloquent,hallucigen,references_hallucigen_evi_thesis}

\onecolumn
\appendix

\section{Hallucination Examples}
\label{sec:hallucination_examples}
Table~\ref{tab:examples} presents examples of hallucinated hypotheses for the paraphrase scenario for each hallucination category.

\begin{table*}[h!]
\centering
\begin{small}
\begin{tabular}{|l|p{6.5cm}|p{6.5cm}|}
\hline
\textbf{Type} & \textbf{Source} & \textbf{Hallucination} \\
\hline\hline
Addition & We struggle with water on a daily basis in the Netherlands - in the polders, the delta where the Meuse, the Rhine and the Scheldt flow into the sea. & 
In the Netherlands, we struggle with water on a daily basis because of the Meuse, Rhine, Scheldt, Noord, Voer and Dieze
\\
\hline
Named-Entity & The fact is that a key omission from the proposals on agricultural policy in Agenda 2000 is a chapter on renewable energy.& Agenda 2030 does not include a chapter on renewable energy.\\
\hline
Number & The European Commission proposes that this information should enter into force within a period of three years from 1 July 1998.& The EU wants this information to enter into force in thirty years.\\
\hline
Conversion & In addition to these losses, there were also significant losses in terms of infrastructures, totalling approximately EUR 15 million.& There were losses in the amount of approximately 15 million dollars.\\
\hline
Date & 	
In 1998, 1 700 000 net jobs were created in Europe, and although I admit that the employment situation is far from ideal, it has improved.& In 1700 there were 1 998 000 net jobs created in Europe.\\
\hline
Gender & 	
Madam President, I am speaking on behalf of our colleague, Mr Francis Decourrière, who drafted one of the motions for a resolution.& One of the motions for a resolution was drafted by Mrs Francis Decourrière.\\
\hline
Pronoun & We have done so: on 5 February we published an extremely detailed press release dealing with the questions you have raised.& We published a press release that dealt with the questions we raised.\\
\hline
Antonym & The population has declined in some 210 of the 280 municipalities in Sweden, mainly in inland central and northern Sweden.& In the majority of Sweden's 280 municipalities, the population has gone up.\\
\hline
Tense & For the latter, the initial birth of several operators is now giving way to the reconcentration of the sector in the hands of a single company. & Several operators have given way to the reconcentration of the sector in the hands of one company.\\
%Den iranska regeringen sa: det iranska kärnkraftsprogrammet förblir fredligt.& Regeringen i Iran förklarade att nationens kärnenergiprogram brukade vara fredligt.\\
\hline
Negation & 	
The draft agenda as drawn up by the Conference of Presidents pursuant to Rule 95 of the Rules of Procedure has been distributed.& The Conference of Presidents hasn't distributed the draft agenda.\\
\hline
Natural & 	
Amendment No 1 in the French version deletes illegal immigration and Amendment No 4 omits the expression 'police authorities'.& The French version excludes the expression'police authorities'.\\
\hline
\cline{2-3}
\hline
\end{tabular}
\end{small}
\caption{Examples of hallucination categories for the paraphrase task.}
\label{tab:examples}
\end{table*}

\newpage

\section{Hallucination Statistics}
\label{sec:category_statistics}
Table \ref{tab:hallucination_cats} presents the frequency of each hallucination category for each language or language pair in the paraphrasing and machine translation hallucination detection scenarios, respectively. The data is first reported by \cite{durlich2024overview}.

\begin{table}[!ht]
    \centering
    \begin{tabular}{cc|rrrrrrrrrrr}
         %& & & & & & & Named & & & & & & \\
         \rotatebox{90}{Language} & \rotatebox{90}{Scenario} %& \multicolumn{10}{c}{Hallucination Category}  \\
         & \rotatebox{90}{Addition} & \rotatebox{90}{Antonym} & \rotatebox{90}{Date} & \rotatebox{90}{Gender} & \rotatebox{90}{Named Entity} & \rotatebox{90}{Negation} & \rotatebox{90}{Number} & \rotatebox{90}{Pronoun} & \rotatebox{90}{Tense} & \rotatebox{90}{Conversion} & \rotatebox{90}{Natural}\\
         \hline
         en & \multirow{2}{*}{PG} & 11 & 16 & 5 & 3 & 9 & 14 & 9 & 11 & 4 & 3 & 33\\
         sv & & 42 & 11 & -- & 3 & 15 & 12 & 9 & 1 & 5 & 1 & 20\\
         \hline
         en-fr & \multirow{4}{*}{MT} & 10 & -- & 24 & -- & 33 & -- & 33 & -- & -- & -- & --\\
         fr-en & & 9 & 13 & 4 & 12 & 12 & 12 & 13 & -- & 12 & 13 & -- \\
         en-de & & 10 & 16 & 14 & -- & 15 & -- & 13 & 16 & -- & -- & 16 \\
         de-en & & 10 & 10 & 7 & 11 & 10 & 10 & 10 & -- & 10 & 11 & 11\\
         
    \end{tabular}
    \caption{Frequency statistics of each hallucination category across the different scenarios and languages.}
    \label{tab:hallucination_cats}
\end{table}

\section{Non-English Prompts}
\label{sec:app_prompts}

Table~\ref{tab:non_English_prompts} presents all non-English prompts used.

\begin{table*}[h!]
    \centering
\scalebox{0.85}{
    \begin{small}
    \begin{tabular}{p{2.5cm} p{12.5cm}} 
    \toprule
    \textbf{Prompt Name} & \textbf{Prompt} \\
    \midrule
     \multicolumn{2}{c}{\textbf{Swedish paraphrase - Swedish prompt}} \\
     \midrule
    Prompt 1 & Givet en mening (src) och två parafrasförslag (hyp1 och hyp2), avgör vilken av de två som är en hallucination av den ursprungliga meningen. En hallucination innebär att hypotesen inte logiskt stöds av källan.\\
    \midrule
    Prompt 2 & Du är en AI-granskare specialiserad på parafrasdetektion.
    
    Din uppgift är följande: Givet en mening (src) och två parafrasförslag (hyp1 och hyp2), avgör vilken av de två som är en hallucination av den ursprungliga meningen. En hallucination innebär att hypotesen inte logiskt stöds av källan.\\
    \midrule
    Prompt 3 & Givet en mening (src) och två parafrasförslag (hyp1 och hyp2), avgör vilken av de två som är en hallucination av den ursprungliga meningen.
    \\
    \midrule
   Prompt 4 & Givet en mening (src) och två parafrasförslag (hyp1 och hyp2), avgör vilken av de två som motsäger den ursprungliga meningen. \\
    \midrule
    Prompt 5 & Givet en mening (src) och två parafrasförslag (hyp1 och hyp2), avgör vilken av de två som stöder den ursprungliga meningen. \\
     \midrule
    Prompt 6 & Givet en mening (src) och två parafrasförslag (hyp1 och hyp2), avgör vilken av de två som är en dålig parafras av den ursprungliga meningen.\\
    \midrule
     \multicolumn{2}{c}{\textbf{Translation fr-en - French prompt}} \\
     \midrule
    Prompt 1 & Étant donné une phrase dans la langue originale (src) et deux hypothèses de traduction de src (hyp1 et hyp2), décide laquelle des hypothèses est une hallucination de src. Une hallucination se manifeste quand l’original ne confirme pas logiquement l’hypothèse.\\
    \midrule
    Prompt 2 & Tu es un réviseur de traduction automatique IA.
    
    Ta tâche est la suivante: Tu reçois une phrase dans la langue originale (src) et deux hypothèses de traduction de src (hyp1 et hyp2). Décide laquelle des hypothèses est une hallucination de src. Une hallucination se manifeste quand l’original ne confirme pas logiquement l’hypothèse.\\
    \midrule
    Prompt 3 & Étant donné une phrase dans la langue originale (src) et deux hypothèses de traduction de src (hyp1 et hyp2), décide laquelle des hypothèses est une hallucination de src. \\
    \midrule
    Prompt 4 &  Étant donné une phrase dans la langue originale (src) et deux hypothèses de traduction de src (hyp1 et hyp2), décide laquelle des hypothèses contredit src.\\
    \midrule
    Prompt 5 & Étant donné une phrase dans la langue originale (src) et deux hypothèses de traduction de src (hyp1 et hyp2), décide laquelle des hypothèses confirme src. \\
    \midrule
    Prompt 6 & Tu es un réviseur IA avec une spécialisation en traduction automatique. Étant donné une phrase dans la langue originale (src) et deux hypothèses de traduction de src (hyp1 et hyp2), décide laquelle des hypothèses est une mauvaise traduction de src.\\
    \midrule
     \multicolumn{2}{c}{\textbf{Translation de-en - German prompt}} \\
     \midrule
    Prompt 1 & Bestimme anhand eines Ausgangssatzes (src) und zweier Übersetzungsvorschläge für src (hyp1 und hyp2), welche dieser zwei Hypothesen halluziniert ist. Eine Halluzination tritt auf, wenn die Hypothese das Original (src) nicht logisch unterstützt.\\
    \midrule
    Prompt 2 & Du bist ein KI-Prüfer für maschinelle Übersetzung.
    
    Deine Aufgabe ist die folgende: Bestimme anhand eines Ausgangssatzes (src) und zweier Übersetzungsvorschläge für src (hyp1 und hyp2), welche dieser zwei Hypothesen halluziniert ist. Eine Halluzination tritt auf, wenn die Hypothese das Original (src) nicht logisch unterstützt.\\
    \midrule
    Prompt 3 & Bestimme anhand eines Ausgangssatzes (src) und zweier Übersetzungsvorschläge für src (hyp1 und hyp2), welche dieser zwei Hypothesen halluziniert ist. \\
    \midrule
    Prompt 4 & Bestimme anhand eines Ausgangssatzes (src) und zweier Übersetzungsvorschläge für src (hyp1 und hyp2), welche dieser zwei Hypothesen src widerspricht.\\
    \midrule
    Prompt 5 & Bestimme anhand eines Ausgangssatzes (src) und zweier Übersetzungsvorschläge für src (hyp1 und hyp2), welche dieser zwei Hypothesen src unterstützt. \\
    \midrule
    Prompt 6 & Du bist ein KI-Prüfer mit Fachkenntnissen in maschineller Übersetzung. Bestimme anhand eines Ausgangssatzes (src) und zweier Übersetzungsvorschläge für src (hyp1 und hyp2), welche dieser zwei Hypothesen eine schlechte Übersetzung von src ist.\\
    \bottomrule
    \end{tabular}
    \end{small}
}%\scalebox
    \caption{Prompt formulations tested in Swedish, French and German.}
    \label{tab:non_English_prompts}
\end{table*}

\newpage

\section{Label Post-Processing}
\label{sec:post-processing}

The tested models usually return one of the two expected labels verbatim ($hyp1$ or $hyp2$), but some models tend to return the label in a different phrasing. For this reason, we first check if the generated model output contains any of these variations: 

\begin{itemize}[topsep=2pt,noitemsep]
   \item ``1'' or ``2''
   \item ``hyp 1'' or ``hyp 2'' (including whitespace)
   \item ``hypotes 1'' or ``hypotes 2''
   \item ``hypothèse 1'' or ``hypothèse 2''
   \item ``hypothese 1'' or ``hypothese 2''
\end{itemize}
If the model output contains only one label (in whatever variation), we extract that as the label. If the generated output contains both labels, we consider the output invalid and return an empty label. If none of the variations above are present, we expand the list of variations to cover the different languages in which the models are prompted:

\begin{itemize}[topsep=2pt,noitemsep]
   \item ``hyp1'' or ``hyp2'' (no whitespace)
   \item ``hypothesis1'' or ``hypothesis12''
   \item ``first or ``second''
   \item ``första or ``andra''
   \item ``erste'' or ``zweite''
   \item ``première/premier'' or ``deuxième''
   \item ``hypotes1'' or ``hypotes2''
   \item ``hypothèse1'' or ``hypothèse2''
   \item ``hypothese1'' or ``hypothese2''
\end{itemize}
As explained in Section~\ref{subsec:experimental_variables}, Prompt 5 is formulated in such a way that the task is reversed; we prompt the model to output a label for the hypothesis that \textit{supports} the source. For this reason, and for this particular prompt only, the label is flipped from $hyp1$ to $hyp2$ and vice versa unless the model produces an empty label (in which case the label is kept as is).

\section{Model repositories}
\label{sec:model-repos}

\begin{table}[htp]
\resizebox{\textwidth}{!}{%
\begin{tabular}{lllr}
\hline
\textbf{Family}          & \textbf{Variant}                   & \textbf{Repository}                                                & \textbf{Version} \\ \hline
\multirow{3}{*}{Llama-3} & \textsc{Meta-Llama-3-8B-Instruct}  & \url{https://huggingface.co/meta-llama/Meta-Llama-3-8B-Instruct}   & 3.0              \\
                         & \textsc{Meta-Llama-3-70B-Instruct} & \url{https://huggingface.co/meta-llama/Meta-Llama-3-70B-Instruct}  & 3.0              \\
                         & \textsc{Meta-Llama-3-70B}          & \url{https://huggingface.co/meta-llama/Meta-Llama-3-70B}           & 3.0              \\ \hline
\multirow{2}{*}{Mixtral} & \textsc{Mixtral-8x7B-Instruct}     & \url{mistralai/Mixtral-8x7B-Instruct-v0.1}                         & v0.1             \\
                         & \textsc{Mixtral-8x22B-Instruct}    & \url{https://huggingface.co/mistralai/Mixtral-8x22B-Instruct-v0.1} & v0.1             \\ \hline
\multirow{2}{*}{EuroLLM} & \textsc{EuroLLM-1.7B}              & \url{https://huggingface.co/utter-project/EuroLLM-1.7B}            & -                \\
                         & \textsc{EuroLLM-1.7B-Instruct}     & \url{https://huggingface.co/utter-project/EuroLLM-1.7B-Instruct}   & -                \\ \hline
\multirow{3}{*}{GPT-SW3} & \textsc{gpt-sw3-20b-instruct}      & \url{https://huggingface.co/AI-Sweden-Models/gpt-sw3-20b-instruct} & -                \\
                         & \textsc{gpt-sw3-20b}               & \url{https://huggingface.co/AI-Sweden-Models/gpt-sw3-20b}          & -                \\
                         & \textsc{gpt-sw3-40b}               & \url{https://huggingface.co/AI-Sweden-Models/gpt-sw3-40b}          & -                \\ \hline
\end{tabular}%
}
\end{table}

\section{NLI Baselines Details}
\label{sec:nli_appendix}

To determine which of the two hypotheses ($hyp1$, $hyp2$) contains a hallucination, we predict  ``entailment`` (E) and ``not\_entailment'' (NE) class scores between the source sentence and each one of the hypotheses. %We then apply the following logic to choose the hallucination:
We then choose the hallucination based on which one or more hypotheses 
\begin{itemize}[topsep=2pt,noitemsep]
    \item If \textbf{E} $>$ \textbf{NE} for one hypothesis and \textbf{E} $<$ \textbf{NE} for the other, we choose the one with \textbf{E} $<$ \textbf{NE}.
    \item If \textbf{E} $>$ \textbf{NE} for both hypotheses, we choose the one with the lowest \textbf{E} score.
    \item If \textbf{E} $<$ \textbf{NE} for both hypotheses, we choose the one with the highest \textbf{NE} score.
\end{itemize}

\section{Compute Environment and Efficiency}
\label{app:compute_env}
The experiments were performed on one cluster \footnote{Details have been suppressed for the sake of anonymity.} equipped with NVidia A100 SXM6 64GB GPUs with a single 32-core Intel Ice Lake CPU. %Nvidia Ampere GPUs with 64GB each.
Model inference is performed sequentially (in other words, without batching) for each sample, using the Accelerate library from Huggingface.\footnote{https://huggingface.co/docs/accelerate/v1.1.0/en/index} Table~\ref{tab:exp_efficiency} presents the number of GPUs used for loading each model, as well as execution time for performing inference on a single model input. 

\begin{table}[htp]
\resizebox{\textwidth}{!}{%
\begin{tabular}{lcc}
\hline
\textbf{Model name} & \textbf{Number of GPUs} & \textbf{Inference time per sample (sec)}\\ \hline
\textsc{Meta-Llama-3-8B-Instruct} & 2 & 7.01 \\
\textsc{Meta-Llama-3-70B} & 4 & 11.44 \\  
\textsc{Meta-Llama-3-70B-Instruct} & 4 & 14.77 \\  
\textsc{Mixtral-8x7B-Instruct}  & 2 &  15.13\\
\textsc{Mixtral-8x22B-Instruct}  & 4 & 23.10 \\
\textsc{EuroLLM-1.7B}  & 1 & 18.34 \\
\textsc{EuroLLM-1.7B-Instruct}  & 1 & 19.94 \\
\textsc{gpt-sw3-20b}  & 1 & 14.45  \\
\textsc{gpt-sw3-20b-instruct}  & 1 & 12.46 \\
\textsc{gpt-sw3-40b} & 3 & 13.02 \\
\hline
\end{tabular}%
}
\caption{Number of GPUs used for loading each model, as well as execution time for performing inference on one input.}
\label{tab:exp_efficiency}
\end{table}

\section{Annotation Guidelines: Paraphrase Hallucinations}
\label{app:ann_guidelines}

\textbf{Task:} Your task is to mark each sentence as hallucination (H) or not hallucination (NH).\\

\noindent
\textbf{Definition of hallucination for this task:} Given a src and a generated hypothesis hyp in the context of paraphrasing, %*,* 
we ask the question: is hyp supported by the src? If yes, then hyp is marked as not hallucination (NH). If no, then hyp is marked as hallucination (H).\\

\noindent
A hypothesis \textit{supports} the source when: 
\begin{itemize}
    \setlength\itemsep{0.01cm}
    \item The overall semantics of the source are preserved, but some minor details are missing 
\end{itemize}

\noindent
A hypothesis \textit{does not support} the source when:
\begin{itemize}
    \setlength\itemsep{0.01cm}
    \item New information, i.e. information that was not present in the source and could not be deduced from the source, is added
    \item It contains nonsensical information (when the source does not)
    \item It misrepresents the semantic relationships in the source (i.e. a bad paraphrase)
\end{itemize}

\noindent
\textbf{Example:} \\

\noindent
\begin{tabular}{ll}
    \toprule
    Src & Stockholm is the capital of Sweden and is located on the East coast \\
    \midrule
    Hyp (NH) & 1) Stockholm, situated on the East coast, serves as the capital of Sweden \\
     & 2) Stockholm is situated on the East coast \\
    \midrule
    Hyp (H) & Stockholm is the capital of Denmark \\
    \bottomrule
\end{tabular}

The annotators for the paraphrase data are the authors of this paper, and all are fluent speakers of English and/or Swedish.

\section{Error Rate by Hallucination Category}
\label{sec:hallucination-type}

\begin{figure}[ht]
    \centering
    \includegraphics[width=0.9\textwidth]{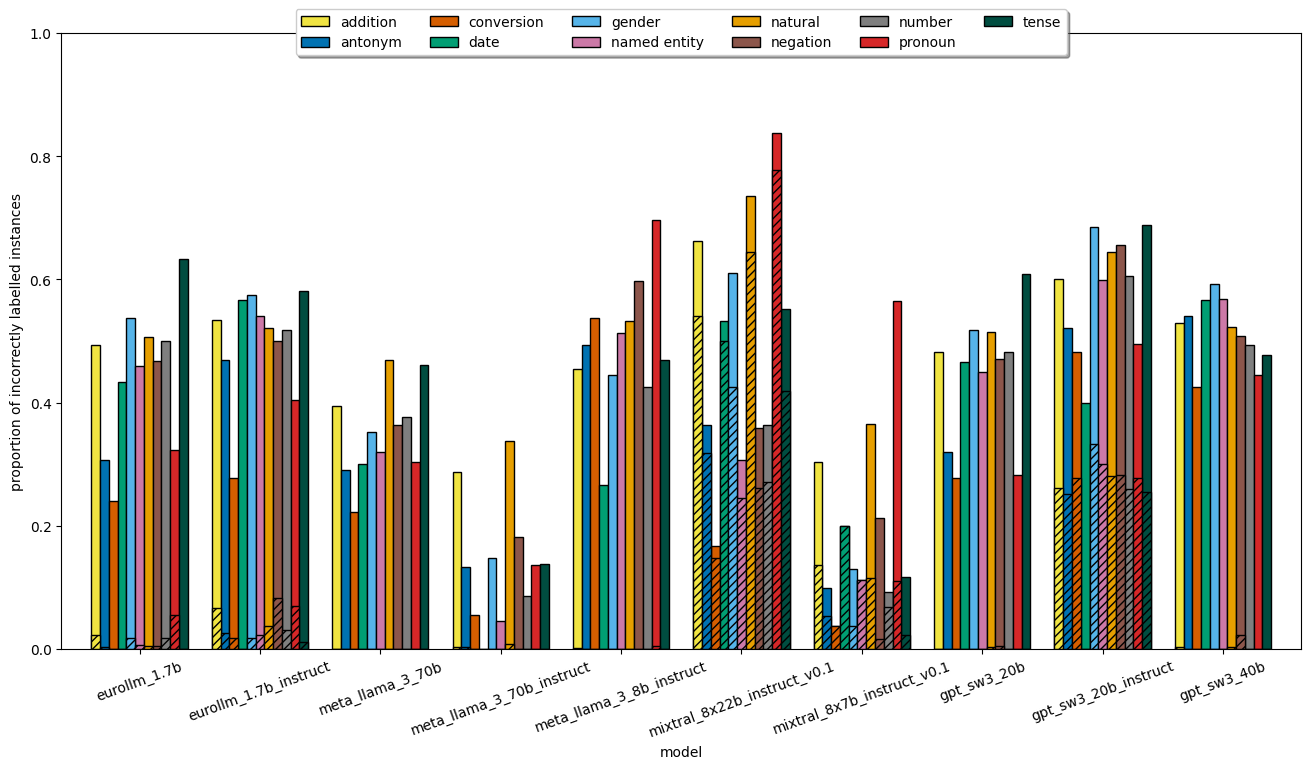}
    \caption{The average proportion of incorrectly labeled source-hyp \textbf{paraphrase pairs} (averaged over all prompts and prompt and data language combinations) filtered by hallucination category. Here, hatch represents the proportion of outputs that were invalid (i.e. falling outside \{$hyp1$, $hyp2$\}).}
    \label{fig:hallucination-type-par}
\end{figure}

\begin{figure}[!b]
    \centering
    \includegraphics[width=0.9\textwidth]{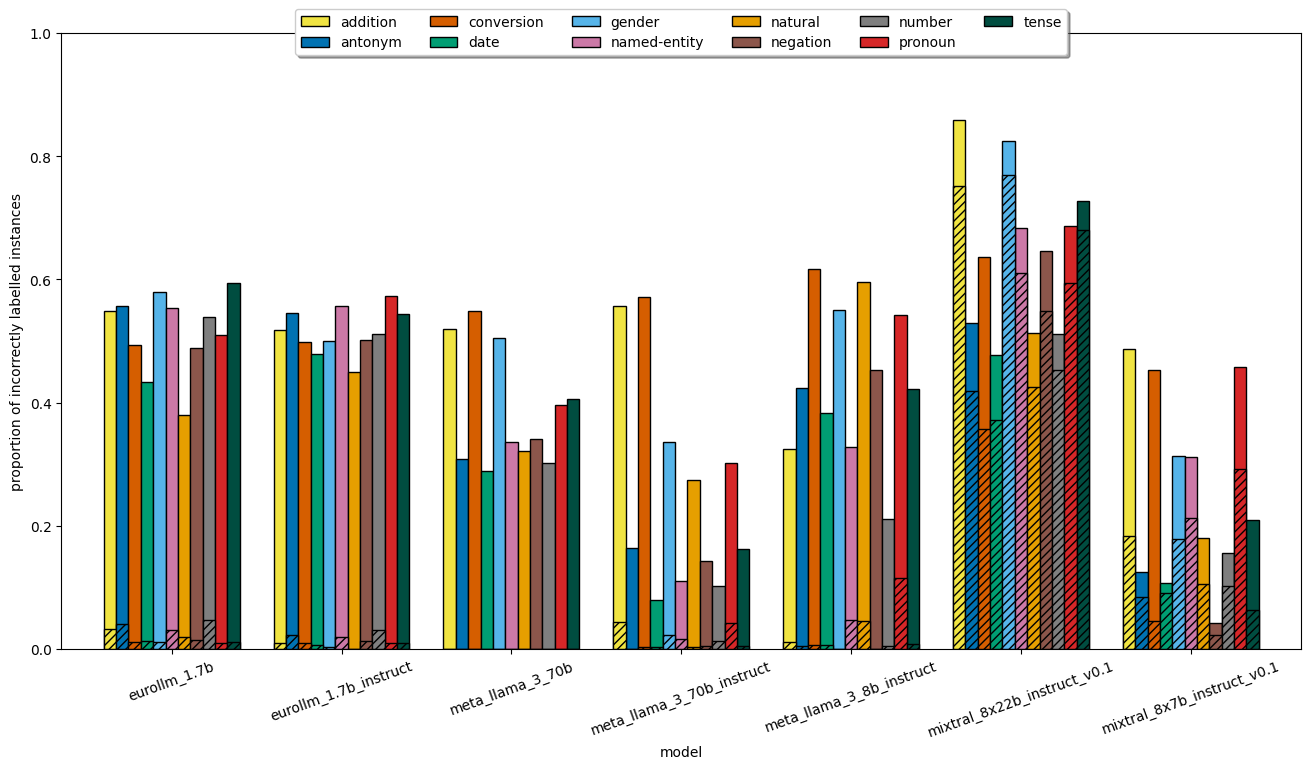}
    \caption{The average proportion of incorrectly labeled source-hyp \textbf{translation pairs} (averaged over all prompts and prompt and data language combinations) filtered by hallucination category. Here, the hatch represents the proportion of outputs that were invalid (i.e. falling outside \{$hyp1$, $hyp2$\}).}
    \label{fig:hallucination-type-trans}
\end{figure}

In Figures \ref{fig:hallucination-type-par} and \ref{fig:hallucination-type-trans}, we provide the error rates (the percentage of incorrect labels) across models averaged over all prompt variations and split by hallucination category. The hatching covers the portion of incorrect labels that were invalid (in other words, they did not correspond to either $hyp1$ or $hyp2$).

\end{document}